\documentclass[11pt]{article}

\usepackage[utf8]{inputenc}
\usepackage[T1]{fontenc}
\usepackage{times} 
\usepackage{graphicx}
\usepackage{subcaption}
\usepackage{xcolor}
\usepackage{amssymb,amsmath,amsthm}
\usepackage{placeins}
\usepackage{hyperref}
\usepackage{pdfpages}
\usepackage{authblk}
\usepackage{caption}
\usepackage{pdfpages}
\usepackage{subcaption}
\usepackage{graphicx}
\usepackage{float}
\usepackage{xcolor}
\usepackage{amssymb,amsmath,amsthm}
\usepackage{placeins} 

\usepackage[margin=1in]{geometry}

\title{DQN Performance with Epsilon Greedy Policies and Prioritized Experience Replay}

\author[1,2]{Daniel Perkins}
\author[1]{Oscar J. Escobar}
\author[1]{Luke Green} 

\affil[1]{Brigham Young University}
\affil[2]{Bredesen Center, University of Tennessee}

\begin{document}

\maketitle

\begin{abstract}
We present a detailed study of Deep Q-Networks in finite environments, emphasizing the impact of epsilon-greedy exploration schedules and prioritized experience replay. Through systematic experimentation, we evaluate how variations in epsilon decay schedules affect learning efficiency, convergence behavior, and reward optimization. We investigate how prioritized experience replay leads to faster convergence and higher returns and show empirical results comparing uniform, no replay, and prioritized strategies across multiple simulations. Our findings illuminate the trade-offs and interactions between exploration strategies and memory management in DQN training, offering practical recommendations for robust reinforcement learning in resource-constrained settings.
\end{abstract}

\section{Background \& Motivation}
\begin{figure}[H]
    \centering
    \includegraphics[scale=0.3]{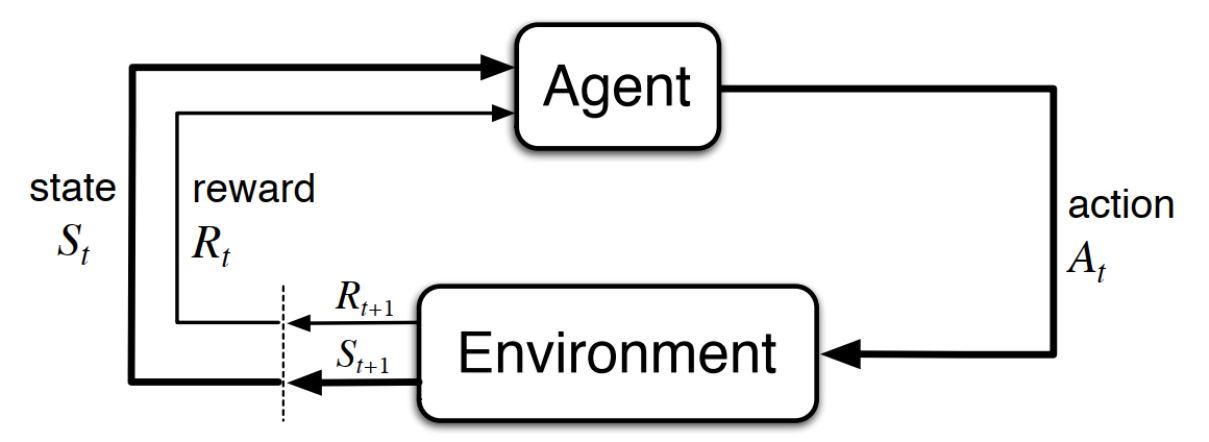}
    \caption{A representation of the RL framework as first introduced by \cite{sutton2018reinforcement}. The agent finding itself in state $S_t$ and having received reward $R_t$ takes action $A_{t+1}$ and transitions into new state $S_{t+1}$ and receives a reward $R_{t+1}$. The cycle then begins again.}
    \label{fig:RL_framework}
\end{figure}

\textit{Reinforcement learning} (RL) is found at the intersection between control theory and machine learning. It was first introduced in the 1950s and is a sequential decision-making process in which an agent learns to perform actions that maximize a specified reward (\cite{prince2023understanding}). The RL framework can be summarized by Fig.~\ref{fig:RL_framework}.

We give definitions of other RL verbiage, such as episode, state, time step, etc., in Appendix~\ref{appendix:defs}. The following definitions are important for this paper:
\begin{itemize}
    \item The return is the discounted sum of future rewards. We denote it as $G_t=\sum_{k=0}^\infty\gamma^kr_{t+k+1}$ where $\gamma\in (0,1]$ is a discount factor that makes rewards closer in time more valuable than those farther away.
    \item The policy is the rule that determines the agent's action at each state. This function may be either deterministic or stochastic. We denote it as $\pi[a|s]$, which is a probability distribution of actions $a$ given the state $s$\footnote{Note that each state $s$ can have a different probability distribution over its own $A_s$.}. Note that it does give an ordering on which actions to take. It only gives a probability for taking an action at a specific state.
    \item The action-value function is the expected future return for a given state-action pair and policy. We denote it as $q_\pi(s,a)=\mathbb{E}[G_t|s_t,\pi]$. It is a way of representing how good the action that the agent takes in a given state is, after considering expected future rewards.
\end{itemize}

Unlike conventional supervised and unsupervised methods, vanilla RL models do not learn from a given dataset and, consequently, have no normal loss function. Instead, they are control algorithms that interact with an environment to complete some task and use the reward as a signal for evaluating performance. The optimization problem of RL is to find an optimal policy that will bring the agent the most future return. Reinforcement learning has the potential to surpass human performance by evaluating and computing a far greater number of future state-action pairs than humans are capable of processing.

\subsection{Q-Learning}
Q-Learning (\cite{watkins1992qlearning}) is an algorithm that learns the optimal policy by calculating state-action values from the direct experiences of the agent's interactions with the environment. That is, Q-learning computes the value of a state-action pair as the agent takes live actions and receives rewards, so that it learns by trial-and-error. In order for Q-learning to converge, there is a need for every possible action and state to be visited infinitely many times \cite{ConvergenceResults}.
This is of course infeasible to calculate for every possible state-action pair in a problem with high dimensional state and action space. 

Specifically, Q-Learning estimates the action-value function by using the reward and the difference in estimated values it has of the current state and an updated estimate. The Q-values are initially set to zero and stored in a matrix. Then, they are iteratively updated with the Bellman equation:
\begin{equation}
    Q_{\text{new}}(s_t,a_t) = Q_{\text{old}}(s_t,a_t) + \alpha \Bigl[ r_t+\gamma\underset{a\in A_{s_{t+1}}}{\text{ max }}Q_{\text{old}}(s_{t+1},a) - Q_{\text{old}}(s_t,a_t) \Bigr]
    \label{eq:qlearning}
\end{equation}
where $\alpha$ is the learning rate and $r_t+\gamma\underset{a\in A_{s_{t+1}}}{\text{ max }}Q_{\text{old}}(s_{t+1},a)$ is the updated approximation of the current state-action pair. 

By learning the value function for each action-value pair, Q-learning is guaranteed to find the optimal policy. It can simply choose the action that maximizes the value at each state the agent finds itself in. When the action space and state space are finite and small, a simple bottom-up dynamic programming implementation will work well (\cite{Watkins}). However, these algorithms are computationally expensive because of their exhaustive search. Therefore, as previously stated, they do not perform well on most real-world tasks which often have large or even infinite action and/or state spaces.

\subsection{RL Challenges} While RL can be quite a good tool for handling sequential decision making, there are a few hurdles that it must overcome.

\subsubsection{Exploration-Exploitation-Trade off}
\label{explore-exploit}
Since Q-learning must learn an optimal policy strictly from experience, the agent has to explore action space to find what actions are good and which are bad. However, we also need the agent to exploit what it has already learned so that the algorithm converges to an optimal policy. Thus, there is a need to properly balance exploration and exploitation if we want the agent to achieve a goal. 

\subsubsection{Credit Assignment Problem}
\label{CAP}
While a reward is a good signal that the agent can use in the immediate sense for judging how good a taken action was in helping it achieve the long term goal, the agent has difficulty determining which actions or sequence of actions actually led to the reward. This is known as the \textit{credit assignment problem} (CAP). 

In order for the agent to accomplish the goal (i.e.\ solving the environment), it must be able to determine which actions actually contribute to solving the environment and which don't. Moreover, due to the sequential nature of RL and indirectly from CAP, how can the agent learn to single out good actions needed to accomplish a task seeing that those come from the sequential decision making? According to Lin, ``if an input pattern has not been presented for quite a while, the
[agent] typically will forget what it has learned for that pattern and thus need to re-learn it
when that pattern is seen again later" \cite{EL_LongJi}.
Thus, for RL problems with bigger state and action spaces, there appears to be a necessity to use previous actions to ``remember" what has been accomplished as well as to disassociate actions. 

\subsection{Deep Reinforcement Learning}

Until recently, Temporal Difference Learning methods like Q-Learning (\cite{watkins1992qlearning}) and SARSA (\cite{RummerySARSA}) were the state of the art in reinforcement learning. Problems with large state spaces were thought to be unsolvable. However, with the advent of deep learning, RL has hit a new frontier. In 2013, fitted Q-learning was introduced (\cite{mnih2013playingatarideepreinforcement}). Instead of storing all possible values in a large matrix, a neural network is used to estimate the action-value function; the Q-values $Q(s_t,a_t)$ are replaced with a neural network $q[s_t,a_t,\phi]$. Since $q[s_t,a_t,\phi]$ should be close to $r_t+\gamma\text{max}_{a\in A}\{q[s_{t+1},a,\phi]\}$, we get the loss function
\begin{equation}
    L(\phi)=(r_t+\gamma\text{max}_{a\in A}\{q[s_{t+1},a,\phi]\}-q[s_t,a_t,\phi])^2
\end{equation}
In theory, performing gradient descent on this function will yield a deep neural network that approximates the true state-value function. However, unlike typical supervised learning algorithms, at each step, the target, $r_t+\gamma\text{max}_{a\in A}\{q[s_{t+1},a,\phi]\}$ is changing at each step. So, to reduce variance, it is common to only alter the target every hundred iterations or so. So, letting $\hat{q}[s_{t+1},a,\phi]$ denote the most recently saved target, this loss function can be written as
\begin{equation}
    L(\phi)=(r_t+\gamma\text{max}_{a\in A}\{\hat{q}[s_{t+1},a,\phi]\}-q[s_t,a_t,\phi])^2
\end{equation}
This new framework has paved the way for solving much more complicated problems. For example because the game Go has a state space with over $10^{170}$ elements. It was once considered to be impossible to develop a RL learning algorithm that can beat the best humans. But, in 2016, a deep RL model beat the \href{https://www.youtube.com/watch?v=WXuK6gekU1Y}{world champion} (\cite{silver2016mastering}).

\subsection{Assumptions \& Objective.} Seeing that there are many stochastic pieces to work with, as given in the definitions, as well as big hurdles that plague RL, we work with a very simplistic model where both our policies and reward are deterministic. Moreover, the environment will have a finite and deterministic nature. Furthermore, to simplify our analysis, we focus on a problem that has a small and finite action and state space as well as having the property that the agent gets a full observation of the state. That is, we work under a Markov Decision Process not a Partially Observable Markov Decision Process. This decreases the number of variables that need to be accounted for.

In particular, we seek to learn how various \textit{epsilon-greedy} policies affect the performance of DQN and if it makes it worse or better than Q-learning. Additionally, we explore the effects of \textit{experience replay}, with decaying \textit{epsilon-greedy} policies, on DQN.

\section{Simulation Environment}

In this study, we utilize the Cart Pole environment from Gymnasium (\cite{gymnasium}) to test our deep reinforcement learning (DRL) algorithm. This environment is well-suited for investigating the mathematical underpinnings of Deep RL, as it features a small action space and relatively straightforward dynamics. The simplicity enables rapid training, which is essential for efficiently testing a wide range of hyperparameter configurations. 
\begin{figure}[H]
\vspace{-2mm}
    \begin{center}
    \includegraphics[scale=0.4]{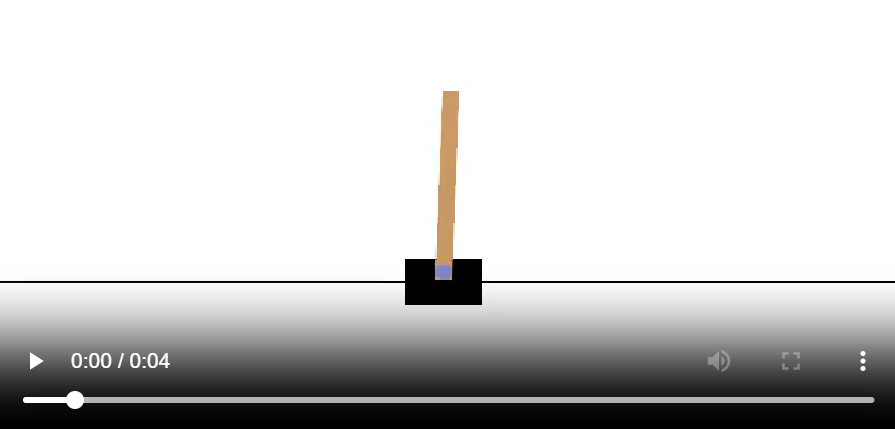}
    \caption{The Cart Pole Environment}
    \label{fig:cartpolt}
    \vspace{-2mm}
    \end{center}
\end{figure}

The Cart Pole Environment presents a setting in which a pole is balanced on a moving cart (Figure \ref{fig:cartpolt}). The goal is to prevent the pole from falling over by adjusting the cart's position. The environment consists of the following:
\begin{itemize}
    \item The state is observed as a four-dimensional vector $s_t=\begin{bmatrix}x_t&v_t&\theta_t&\omega_t\end{bmatrix}$ where $x_t\in [-4.8,4.8]$ and $v_t\in(-\infty,\infty)$ denote the cart's position and velocity, respectively, and $\theta_t\in(-0.418,0.418)$ and $\omega_t\in(-\infty,\infty)$ represent the pole's angle and angular velocity.
    \item The initial state $s_0$ is an observation where each element of $s_0$ is sampled uniformly from $(-0.05, 0.05)$.
    \item The action space $A=\{0,1\}$ is the same at each state. $0$ denotes the action of moving to the left and $1$ is the action of moving to the right.
    \item At each time step, the reward $r(s_t,a_t)$ is $0$ if the pole falls and $1$ otherwise.
    \item The simulation ends if the pole angle $\theta_t$ is greater than $\pm 0.2095$, if the cart position $x_t$ is greater than $\pm 2.4$, or if the episode length $t$ is greater than $500$.
\end{itemize}

\section{Q-Learning}
\label{q}
\begin{figure}[H]
    \vspace{-2mm}
    \centering
    \includegraphics[scale=0.4]{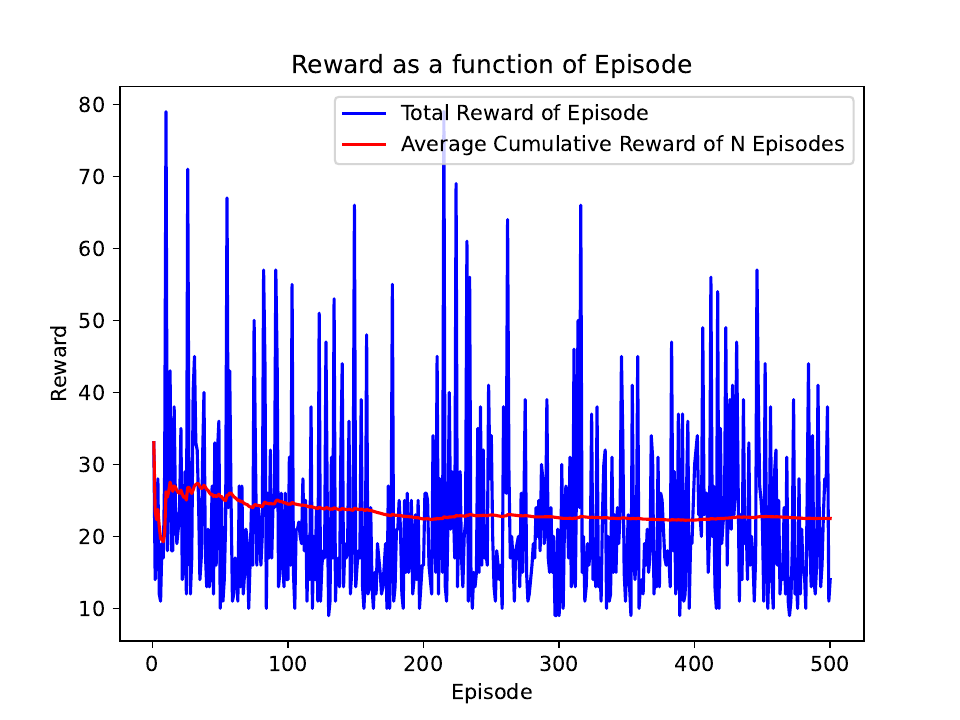}
    \caption{The rewards obtained using an exponential decay $0.9999^t$ for $\varepsilon$-greedy in normal Q-learning. Compare to Fig.~\ref{fig:exponentialdecay}.}
    \label{fig:qlearn_lin_dec}
    \vspace{-2mm}
\end{figure}

We first solve this environment using normal Q-learning in\ \eqref{eq:qlearning}. We discretized the state space by taking a certain number of values for each of the components of $s_t$. We employed a linear decay for the epsilon-greedy algorithm (explained in \ref{epsilon-greedy}). Our results are shown Fig.~\ref{fig:qlearn_lin_dec}.

As shown, the results are fairly good with normal Q-learning. The agent is able to accomplish the task of balancing the pole, and with more episodes, it performs better on average. Note that due to the stochastic nature of our policy the rewards are noisy as shown in the graph. Notice also how the average reward (red) somewhat increases at times but stays overall stagnant and even decreases at times.

\section{Deep Q-Network: Epsilon Greedy Algorithm/Policy}
\label{epsilon-greedy}
Although the results in Section \ref{q} are promising, they remain inherently constrained to relatively small state spaces. Q-learning struggles to generalize to more complex tasks, motivating the use of neural networks to approximate Q-values in Deep Q-Networks (DQNs). However, due to the vast size of complex action spaces, DQNs are unable to encounter all possible states during training and can only learn from those observed. Therefore, it is essential to develop an algorithm that effectively balances exploration of new states with exploitation of the knowledge gained from prior experience.

To balance exploration and exploitation (Section \ref{explore-exploit}), DRL commonly uses the \textit{epsilon-greedy algorithm}, which specifies a value $\varepsilon\in[0,1]$ that signifies the rate at which the agent must explore. By drawing some random value from $c\sim\text{Uniform}[0,1]$, we can \textit{explore} the action space when $c<\varepsilon$ or \textit{exploit} by taking the best action when $c\geq\varepsilon$. This strategy enables the agent to explore alternative actions that could potentially yield higher rewards than the current policy.  

Setting $\varepsilon$ too high greatly increases the temporal complexity of the algorithm. But, setting it too low reduces the probability that it finds the optimal policy. It is common to let $\varepsilon$ decay overtime so that the model learns more in the beginning and exploits what it learns later on. In our paper, we investigate the effects of various schedules for $\varepsilon$.

\subsection{Theory}
\label{theory}
\cite{ConvergenceResults} gives proofs for guaranteed convergence of an algorithm similar to Q-learning, called \textit{SARSA(0)}\footnote{SARSA(0) is given by $Q_{\text{new}(s_t,a_t)}=Q_{\text{old}} (s_{t}, a_{t}) + \alpha\Bigl[r_t + \gamma Q_{\text{old}} (s_{t+1}, a_{t+1}) - Q_{\text{old}}(s_t,a_t)\Bigr]$. It uses the tuple $(s_t, a_t, r_{t+1}, s_{t+1}, a_{t+1})$, hence the name. Compare to \eqref{eq:qlearning}.}, employing different behavioral/learning policies (Appendix ~\ref{appendix:defs}). They discuss how these concepts apply to Q-learning. In particular, they state that when using decaying behavioral policies, actions may converge to the optimal actions but at the cost of losing ability to adapt as the value decays. Moreover, they identify two crucial properties that decaying policies should have: 1) ``each action is visited infinitely often in every state that is visited infinitely often", and 2) ``in the limit, the learning policy is greedy with respect to the Q-value function with probability 1".

\cite{ConvergenceResults} represents this mathematically by letting $t_s(i)$ denote the time step at which the $i^{\text{th}}$ visit to state $s$ occurs and considering some action $a$. To visit ever state infinitely often, they show that probability of executing action $a$ must satisfy $\sum_{i=1}^\infty \mathbb{P}(a=a_t|s_t=s, t_s(i)=t)=\infty$\cite{ConvergenceResults}. This of course is affected by $\varepsilon$-greedy since it takes actions by comparing a randomly sampled value $c\sim\text{Unif}[0,1]$ to the current value $\varepsilon$. Thus, careful consideration must be given to how we decay $\varepsilon$ to obtain optimal results. 

\subsection{Exponential Decay} One of the most common schedules for $\varepsilon$ is exponential decay. At each iteration $t$, we set $\varepsilon = \beta^t$ for some $\beta \in [0,1]$. This simple formula gradually reduces the exploration rate as the agent gains more experience, allowing it to increasingly exploit the learned policy while still exploring in the early stages. By adjusting $\beta$, we can control the rate at which exploration diminishes, balancing the trade-off between discovering new strategies and refining the current policy.

\begin{figure}[H]
    \begin{center}
    \includegraphics[scale=0.4]{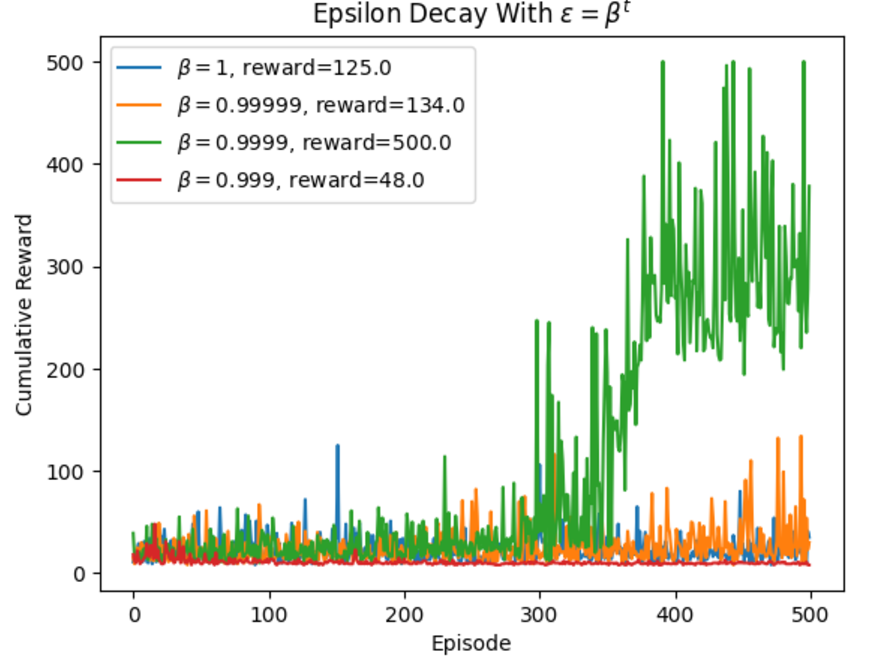}
    \caption{Cumulative reward with exponential epsilon decay for various values of $\beta$}
    \label{fig:exponentialdecay}
    \end{center}
\end{figure}
Figure~\ref{fig:exponentialdecay} shows the empirical results for the deep Q-learning model with an exponential decay schedule for the epsilon-greedy algorithm. When $\beta$ was too large, the agent selected random actions too frequently, resulting in lower rewards. Conversely, when $\beta$ was too small, the model was able to learn effectively only in the early iterations, ultimately getting stuck in a suboptimal solution. We found that for the Cart Pole problem, the optimal value of $\beta$ was 0.9999, as it produced the highest cumulative reward.

\subsection{Other Decaying Epsilon Schedules}

\begin{figure}[H]
    \centering
    \begin{subfigure}[b]{0.45\textwidth}
        \centering
        \includegraphics[width=\linewidth]{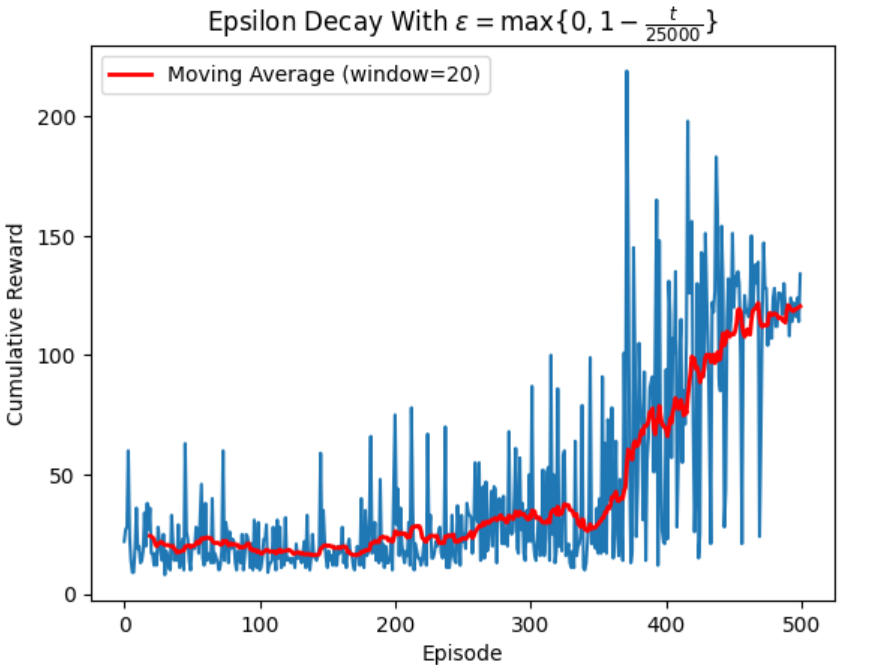}
        \caption{Linear epsilon decay with the rule $\varepsilon=\text{max}(0,1-\frac{t}{25000})$}
        \label{fig:subfig1}
    \end{subfigure}
    \hfill
    \begin{subfigure}[b]{0.45\textwidth}
        \centering
        \includegraphics[width=\linewidth]{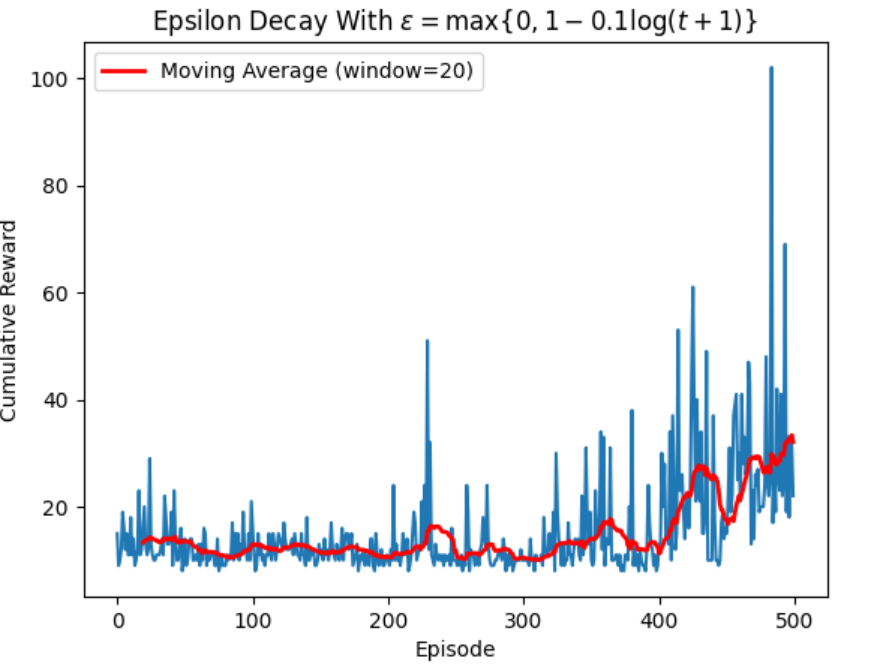}
        \caption{Logarithmic epsilon decay with the rule $\varepsilon=\text{max}(0,1-\frac{1}{10}\log(t+1))$}
        \label{fig:subfig2}
    \end{subfigure}

    \vspace{1em}
    
    \begin{subfigure}[b]{0.45\textwidth}
        \centering
        \includegraphics[width=\linewidth]{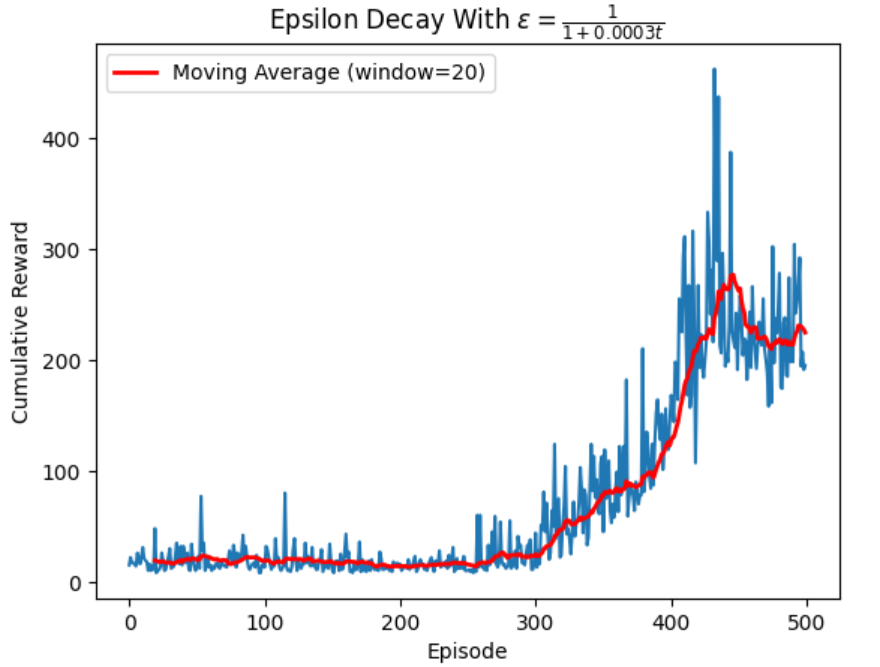}
        \caption{Inverse epsilon decay with the rule $\varepsilon=\frac{1}{1+\frac{3}{1000}t}$}
        \label{fig:subfig3}
    \end{subfigure}
    \hfill
    \begin{subfigure}[b]{0.45\textwidth}
        \centering
        \includegraphics[width=\linewidth]{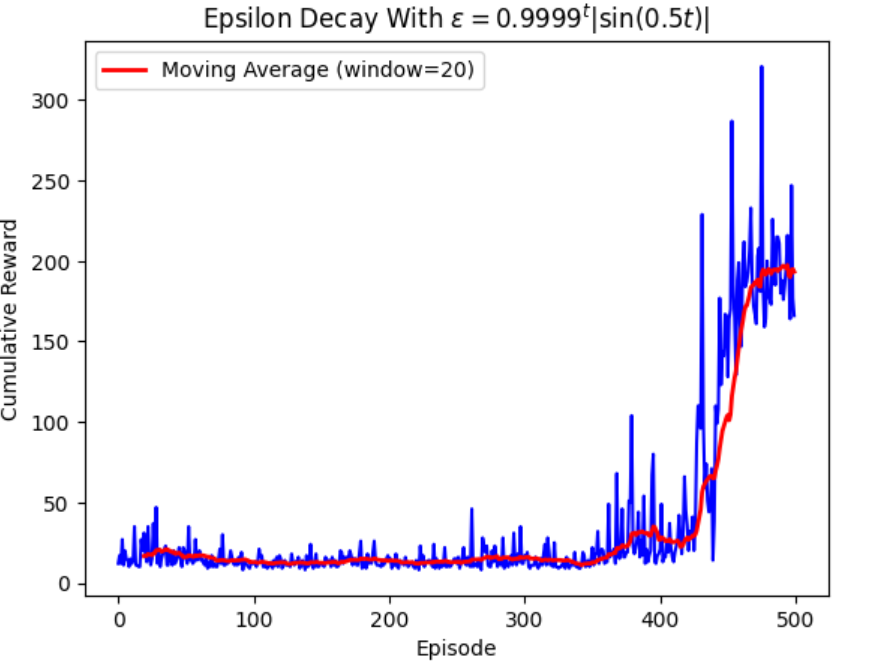}
        \caption{Sinusoidal epsilon decay with the rule $\varepsilon=0.9999^t|\sin(\frac{1}{2}t)|$}
        \label{fig:subfig4}
    \end{subfigure}
    
    \caption{Cumulative reward with various epsilon decay schedules}
    \label{fig:decaygrid}
\end{figure}

We show results of our investigations for various other epsilon decaying schedules in Figure~\ref{fig:decaygrid}.
The linear (\ref{fig:subfig1}) and logarithmic (\ref{fig:subfig2}) schedules did not obtain rewards as high as the inverse (\ref{fig:subfig3}) and sinusoidal decay (\ref{fig:subfig4}) schedules. This suggests that the DRL is more likely to get better results if the decay of $\varepsilon$ is rapid (super-linear). The inverse decay schedule has the highest average reward empirically. However, this does not necessarily mean that it is the optimal schedule. All of the schedules we tested were noisy, differing substantially in each run.

Each schedule demands considerable hyperparameter tuning to determine the ideal rate of decay. Consequently, we conclude that the optimal selection of the $\varepsilon$-decay algorithm and its associated hyperparameters is highly task-dependent.

\section{Deep Q-Network: Experience Replay}
\label{experience_replay}
As mentioned in Section \ \ref{CAP}, two big hurdles in RL are disassociating actions from their sequential nature and then remembering what is already known. \cite{EL_LongJi} introduces \textit{experience replay}, a DQN algorithm that use of a replay buffer, $\mathcal{D}$ of some fixed size $D$, that stores the tuple $e_t=(s_t, a_t, r_{t+1}, s_{t+1})$, as the agent steps through various time-steps $t$. The aim of this was to give a concise summary of the experience or lesson the agent went through at any time $t$. 

In the simple implementation, the DQN algorithm uniformly at random samples some batch $E=\{e_i\}_{i=1}^N$ comprised experiences and then uses the behavioral/policy network with $(s_i, a_i)\in e_i$ in order to compute the current q-value $q(s_i, a_i)$ for each $e_i\in E$. The optimal/target q-value for $(s_i, a_i)$ is computed by using the formula $q^*(s_i, a_i) = r_{i+1} + \gamma \underset{a_{i+1}\in A_{s_{i+1}}}{\max}\text{ }q(s_{i+1}, a_{i+1})$, where $q$ is \textit{target} network (Appendix ~\ref{appendix:defs}). This is subtracted from the predicted value to get the loss. Gradient descent is used to minimize this loss, encouraging the model to converge to optimal q-values. 

Experience replay allows the agent to ``remember" previous actions and, most importantly, learn how to break the correlation among actions due to the sequential nature from which it experiences them. Moreover, as shown in \cite{fedus2020revisitingfundamentalsexperiencereplay}, experience replay improves stability and leads to faster convergence.
\begin{figure}[H]

    \centering
    \includegraphics[scale=0.4]{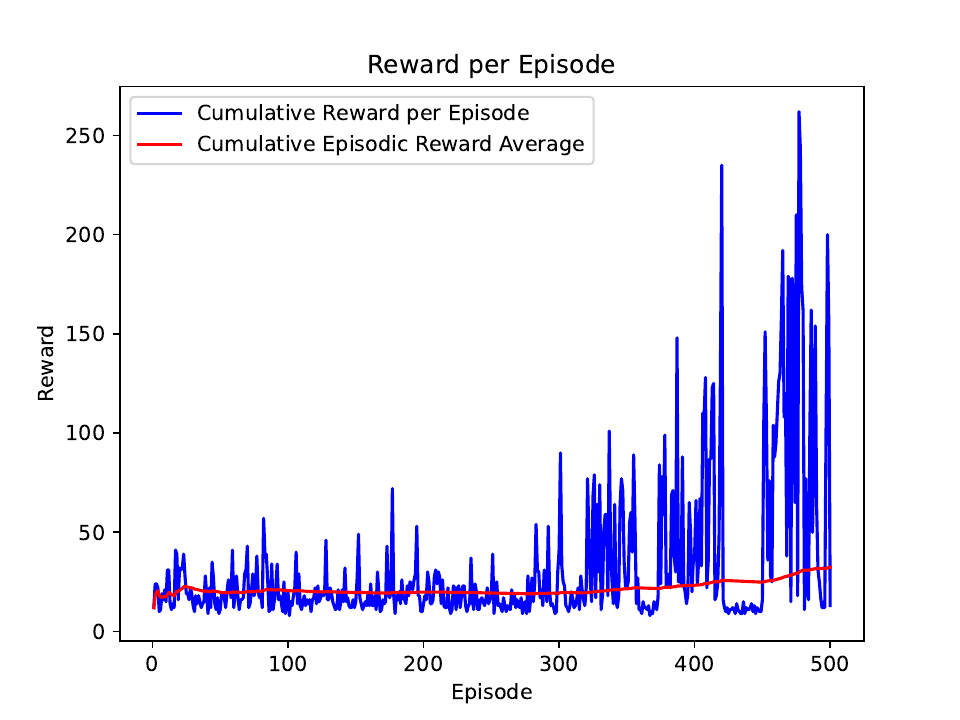}
    \caption{The cumulative reward obtained at each episode using a DQN without any experience replay. Compare to Figures~\ref{fig:qlearn_lin_dec} and~\ref{fig:perexponentialdecay} . Notice how the results for the latter episodes are significantly higher than those obtained by Q-learning but only about half those given by prioritized experience replay.}
    \label{fig:dqn_no_experience}
\end{figure}

\subsection{No Experience Replay.} We plot the results of DQN without any experience replay in Fig.~\ref{fig:dqn_no_experience}. Despite not being able to perform the disassociation or remembrance, the use of a 5-layer linear network comprised of ReLu activation functions and 2 hidden layers with a width of 8 neurons, allowed us to significantly improve, at times, the reward obtained by normal Q-learning. However, due to the stochastic nature of our policies, there were times where the network only did slightly better than Q-learning. 

This can be explained by the fact that the hidden layers can compute features that the normal Q-learning function is not able to experience or compute. Moreover, the layers can adapt to a continuous state-space and do not need discretization, which is a major weakness of Q-learning.

\subsection{Prioritized Experience Replay}
Prioritized Experience Replay (PER) was first introduced by (\cite{schaul2016prioritizedexperiencereplay}) in 2016 as a way to further improve (Uniform) Experience Replay by replaying experiences from which we can learn the most more frequently. This is measured by the Temporal Difference Error:
\begin{equation}
    | r_t+\gamma\text{max}_{a\in A}\{q[s_{t+1},a,\phi]\}-q[s_t,a_t,\phi] |.
\end{equation}
The priority $p_i$ of the $i^{\text{th}}$ experience is equivalent to its TD-error. We then assign a probability to each experience in the replay buffer based on its priority:
\begin{equation}
    P(i)= \frac{p_i^\alpha}{\sum_k p_k^\alpha}.
\end{equation}
The hyperparameter $\alpha$ controls how much weight we are giving to the probabilities. We only recalculate the TD-errors of sampled experiences in order to reduce the computation time. By replaying the experiences with high TD-error more frequently, we introduce bias into our model. In order to counteract this, every time an experience is replayed, we must decrease its effect on the training updates. We do this by weighting the errors $\delta_i$ as $w_i\delta_i$, where $w_i$ is defined as
\begin{equation}
    w_i= \Biggl(\frac{1}{N} \cdot \frac{1}{P(i)} \Biggr)^\beta\cdot\frac{1}{\underset{i}{\max}\text{ } \space w_i},
\end{equation}
$N$ being the size of the replay buffer and $\beta$ emphasizing how much affect the weight should have. When implemented, $\beta$ should start at $\beta_0$ and increase to $1$. The best results came by initializing $\beta_0 = 0.4$. 

\begin{figure}[H]
    \begin{center}
    \includegraphics[scale=0.4]{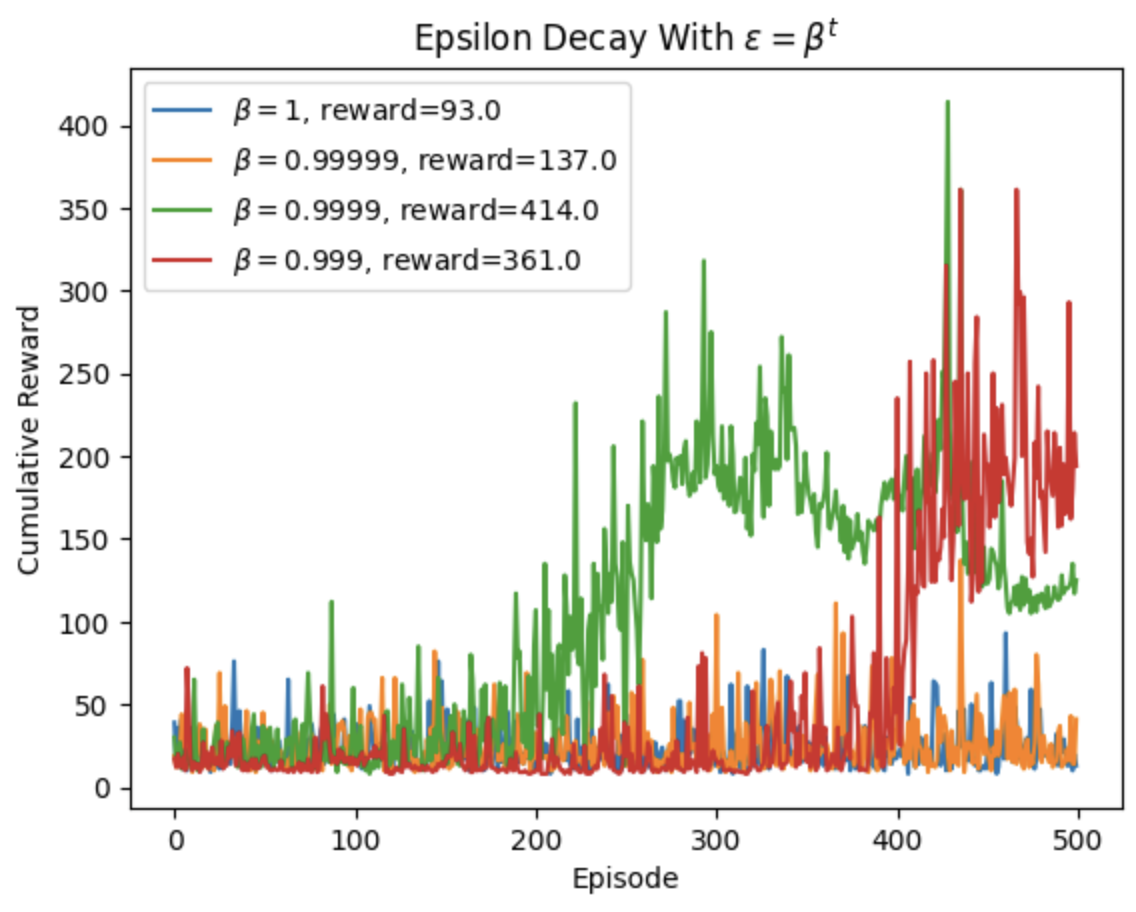}
    \caption{Cumulative reward with Prioritized Experience Replay Buffer and exponential epsilon decay for various values of $\beta$}
    \label{fig:perexponentialdecay}
    \end{center}
\end{figure}

Figure~\ref{fig:perexponentialdecay} shows the performance of exponential epsilon decay with Prioritized Experience Replay. Overall, the results were very similar to those found in Figure~\ref{fig:exponentialdecay}, which used normal/uniform experience replay. For $\beta = 0.9999$, the cumulative reward exceeds 200 before reaching its 300th episode, demonstrating how PER can learn more efficiently. Likewise, we see that for $\beta = 0.999$, performance exceeded a cumulative reward of 200 after 400 episodes. Overall, PER was slightly more efficient, though, no more accurate than uniform experience replay. In fact, for $\beta = 0.9999$, the cumulative reward was about 100 points short of that reached by the uniform replay experience. As stated before, performance varied widely on each run. Therefore, it is difficult to really see how these two implementations compare.

\begin{figure}[H]
    \centering
    \begin{subfigure}[b]{0.45\textwidth}
        \centering
        \includegraphics[width=\linewidth]{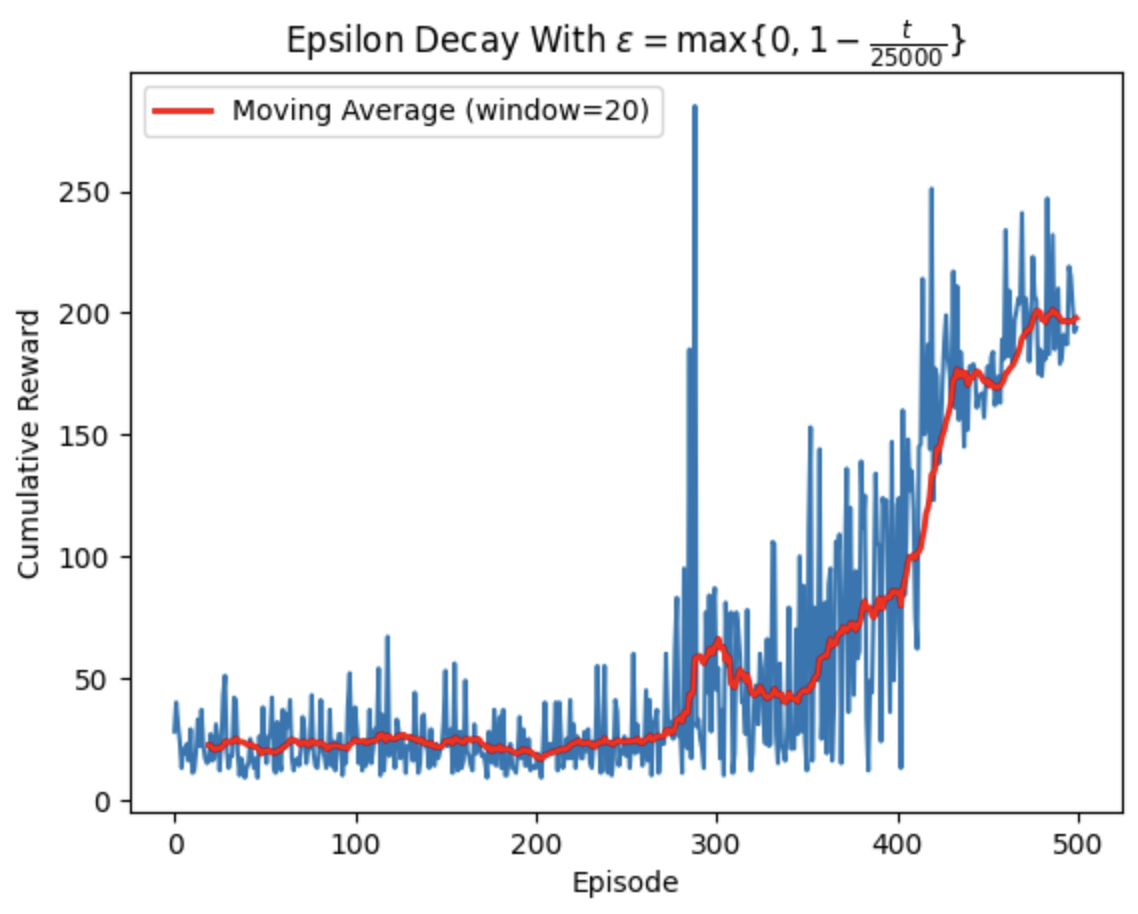}
        \caption{PER and Linear epsilon decay with the rule $\varepsilon=\text{max}(0,1-\frac{t}{25000})$}
        \label{fig:persubfig1}
    \end{subfigure}
    \hfill
    \begin{subfigure}[b]{0.45\textwidth}
        \centering
        \includegraphics[width=\linewidth]{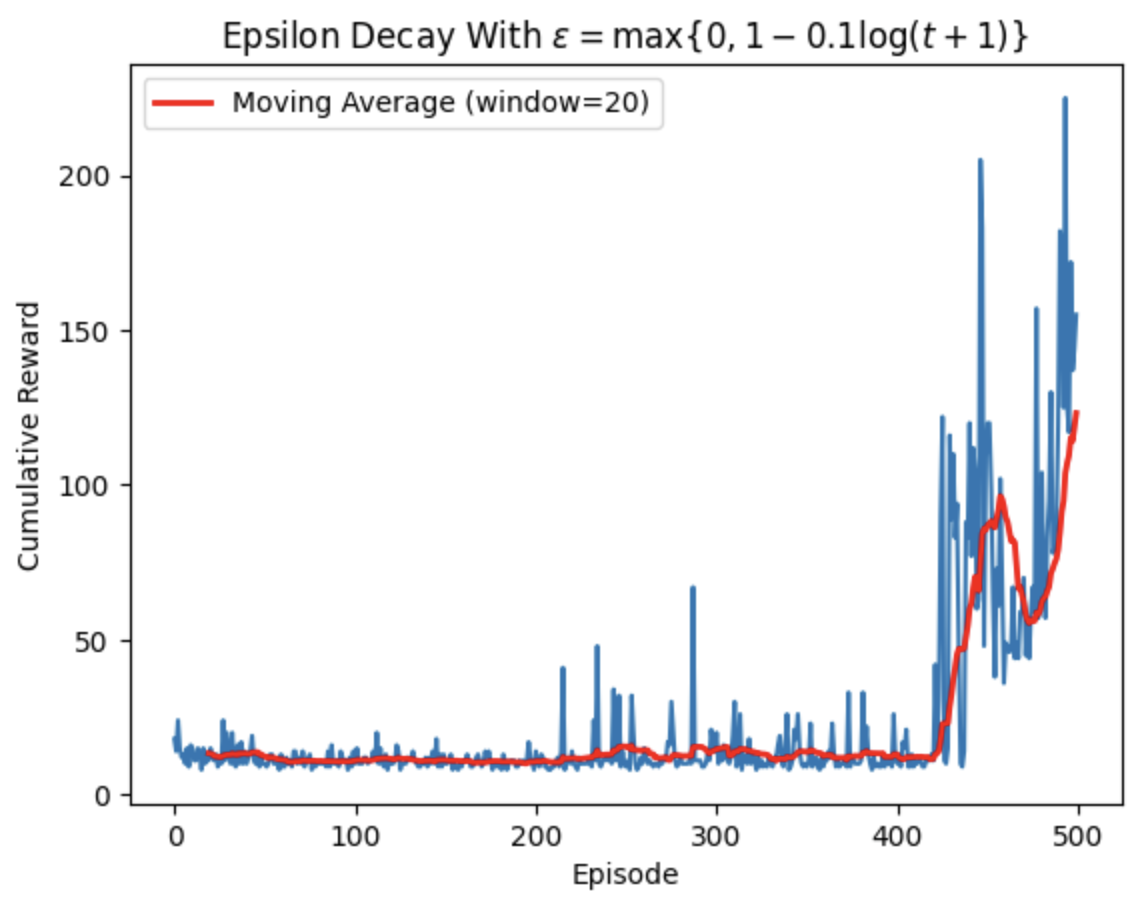}
        \caption{PER and Log epsilon decay with the rule $\varepsilon=\text{max}(0,1-\frac{1}{10}\log(t+1))$}
        \label{fig:persubfig2}
    \end{subfigure}

    \vspace{1em}
    
    \begin{subfigure}[b]{0.45\textwidth}
        \centering
        \includegraphics[width=\linewidth]{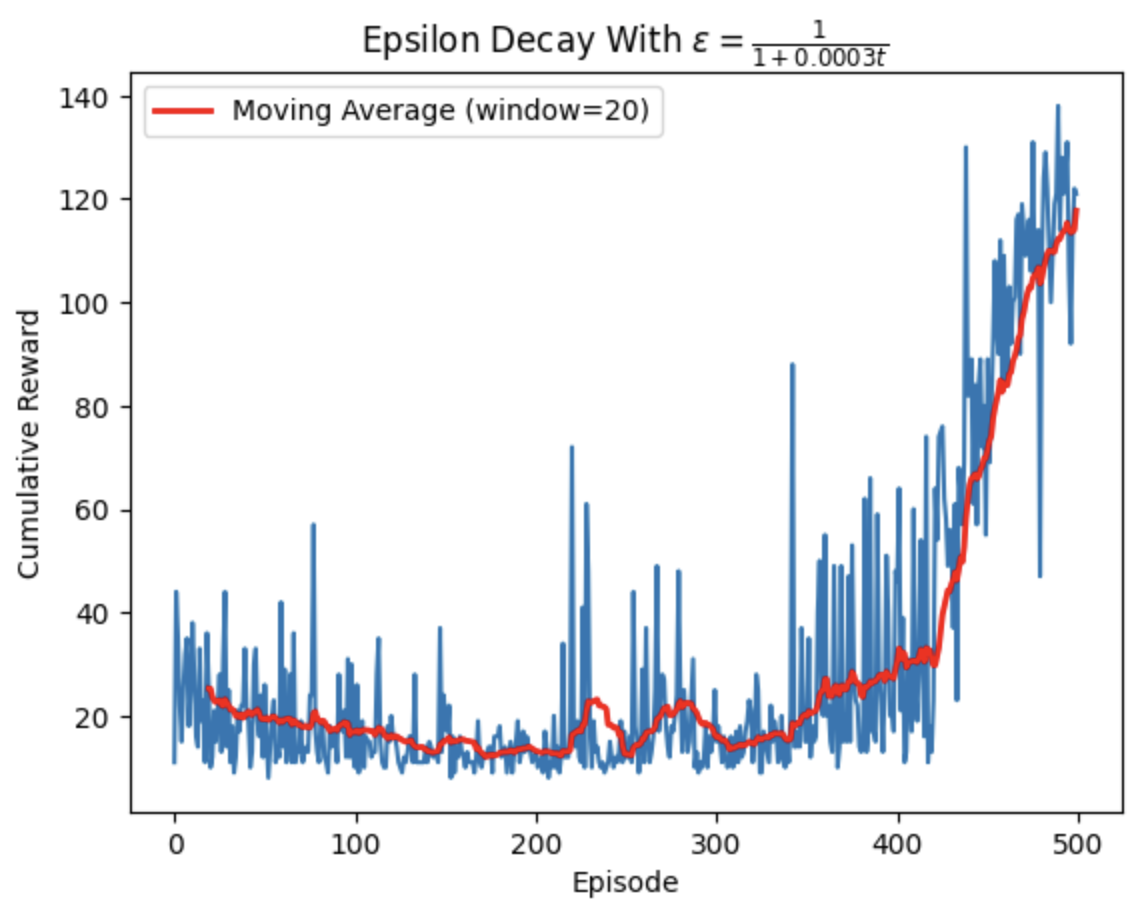}
        \caption{PER and Inverse epsilon decay with the rule $\varepsilon=\frac{1}{1+\frac{3}{1000}t}$}
        \label{fig:persubfig3}
    \end{subfigure}
    \hfill
    \begin{subfigure}[b]{0.45\textwidth}
        \centering
        \includegraphics[width=\linewidth]{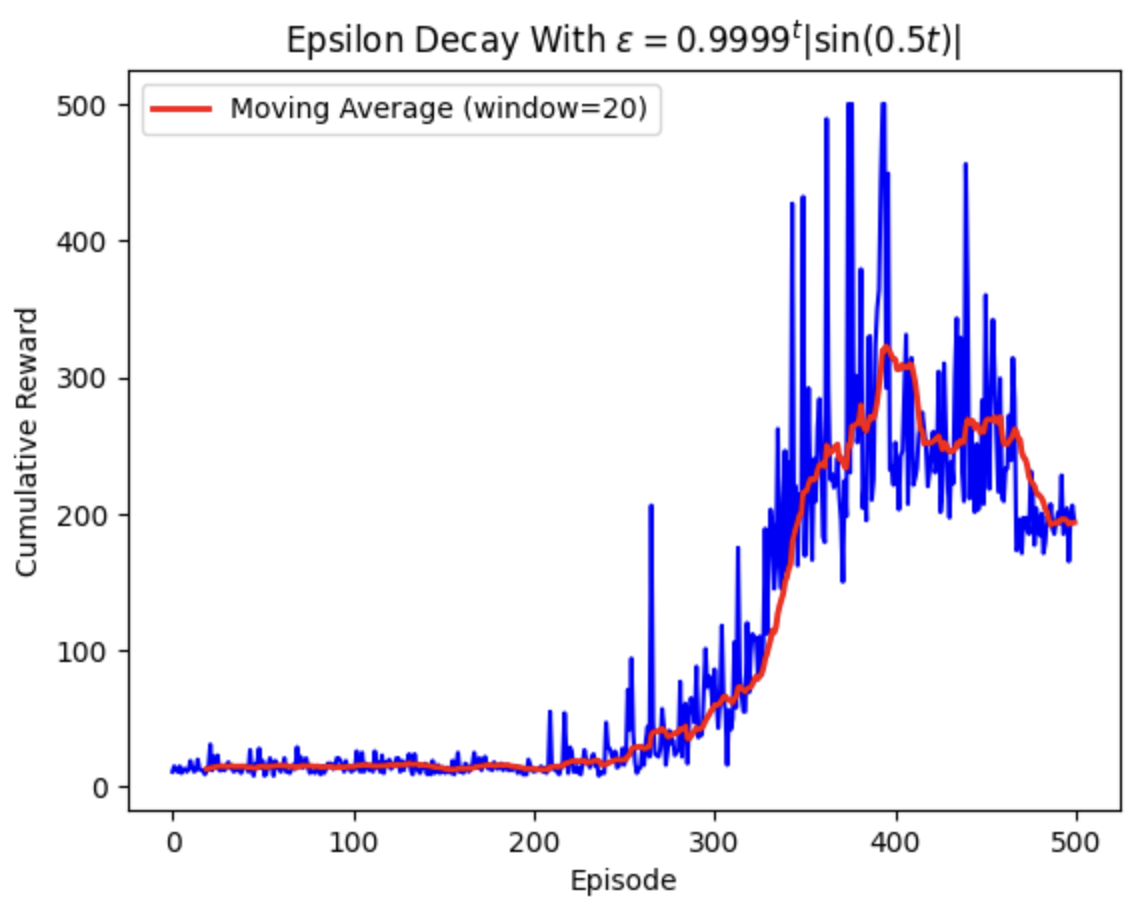}
        \caption{PER and Sinusoidal epsilon decay with the rule $\varepsilon=0.9999^t|\sin(\frac{1}{2}t)|$}
        \label{fig:persubfig4}
    \end{subfigure}
    
    \caption{Cumulative reward with PER and various epsilon decay schedules}
    \label{fig:perdecaygrid}
\end{figure}

The results were similar for the other epsilon decay schedules reinforced with PER. \ref{fig:persubfig1}, \ref{fig:persubfig2}, and \ref{fig:persubfig4} show that each model reached a cumulative reward about 100 points higher than their respective models seen in \ref{fig:subfig1}, \ref{fig:subfig2}, and \ref{fig:subfig4}. However, inverse epsilon decay, which had the best performance with a uniform experience replay (reaching nearly 300 points), performed dreadfully with PER as seen in \ref{fig:persubfig3}. Overall, the Deep Q Network was able to learn in fewer episodes, but the trained model didn't always perform as well as it did with traditional uniform experience replay. In addition, the runtime for PER averaged about 2 minutes, while uniform experience replay averaged 20 seconds. So, even though PER learned in fewer episodes, the runtime was still longer overall. 

We hypothesize that PER is more effective in problems involving more complex environments. Since the CartPole environment exhibited predictable dynamics and a small state space, uniform experience replay was sufficient. However, in more challenging domains with high-dimensional observations and stochastic transitions, PER is necessary because it can significantly improve sample efficiency by prioritizing the most informative experiences.

\section{Conclusion}
Overall, our results demonstrate that integrating neural networks into Q-learning markedly enhances performance, reducing the number of episodes required to achieve high cumulative rewards. As shown in Fig.~\ref{fig:qlearn_10k}, standard Q-learning is inefficient in larger state spaces. Deep Q-Networks (DQNs) extend this capability by enabling more effective generalization across reinforcement learning problems, though their success strongly depends on properly balancing exploration and exploitation.

Our findings underscore the importance of extensive exploration during the initial training stages, followed by focused exploitation to refine performance. The most effective epsilon-decay strategies for the CartPole environment followed super-linear decay schedules, supporting this balance. Additionally, prioritized experience replay can improve both the stability and learning speed of DQNs in more complex environments. Nevertheless, achieving optimal performance still relies on careful hyperparameter tuning tailored to the specific dynamics of each task.

\section*{Acknowledgments}

We thank Dr. Jared Whitehead, and the Math Department of Brigham Young University for supporting this research project.

\bibliographystyle{plain}
\bibliography{refs}

\newpage
\appendix

\section{Reinforcement Learning Definitions}
\label{appendix:defs}
\begin{itemize}
    \item The agent is the decision maker that learns a sequence of actions to perform for a given task.

    \item The action space, is the set of all possible actions the agent can perform. We denote this as $A$, where $a_t\in A$ is a specific action at time $t$. $A_s$, called the \textit{set of allowable actions} or \textit{action set}, denotes set of actions from $A$ that an agent can take during a specific state. The action set can be different for each state.
    
    \item The behavioral or learning policy is the policy in charge of selecting actions based on the current state and obtained reward (e.g. $\varepsilon$-greedy in our study). In DQN, we call the behavioral policy the \textit{behavioral network} or \textit{policy network} since we employ a neural network to compute it.
    
    \item The environment is the world in which the agent is located. It is usually stochastic in nature. 

    \item An observation is the agent's measurement or perception of the state of the environment. Thus, the observation need not be the full state (i.e. MDP vs POMDP).
    
    \item The reward is a stochastic or deterministic function that assigns real values to states and actions in order to signal an immediate outcome.  The reward helps the agent measure in the immediate sense the value of a specific state-action pair.

    \item The state space, is the set of all possible representations of the environment. We denote this as $S$, where $s_t\in S$ is a specific state at time $t$.

    \item The target policy is the policy in charge of updating the actual q-values. In DQN and Q-learning, it always exploits (i.e. the max over the actions). The network, in DQN, in charge of computing this policy is termed the \textit{target network}.

    \item A time step $t$ is the smallest discrete unit of time where the agent interacts with the environment once. This interaction typically comprises a cycle of one state, one action, and one reward. An \textit{episode} is a finite sequence of time steps that starts at some \textit{initial state}, $s_0$ (i.e. the state at $t = 0$), and ends at some \textit{terminal state}, $s_T$ . The \textit{terminal state} is the final state representing the end of an episode and can be a maximum number of time steps or some other desired state. Note that in this latter case, the \textit{time of termination}, $T$, can be different for each episode as well as the fact that each episode can have a different terminal state.
    
\end{itemize}

\section{Additional Results}
\label{appendix:Additional Graphs}
In this section we give additional graphs that further visualize our results. 
\vspace{-0.625cm}

\begin{figure}[H]
    \centering
    \includegraphics[scale=0.6]{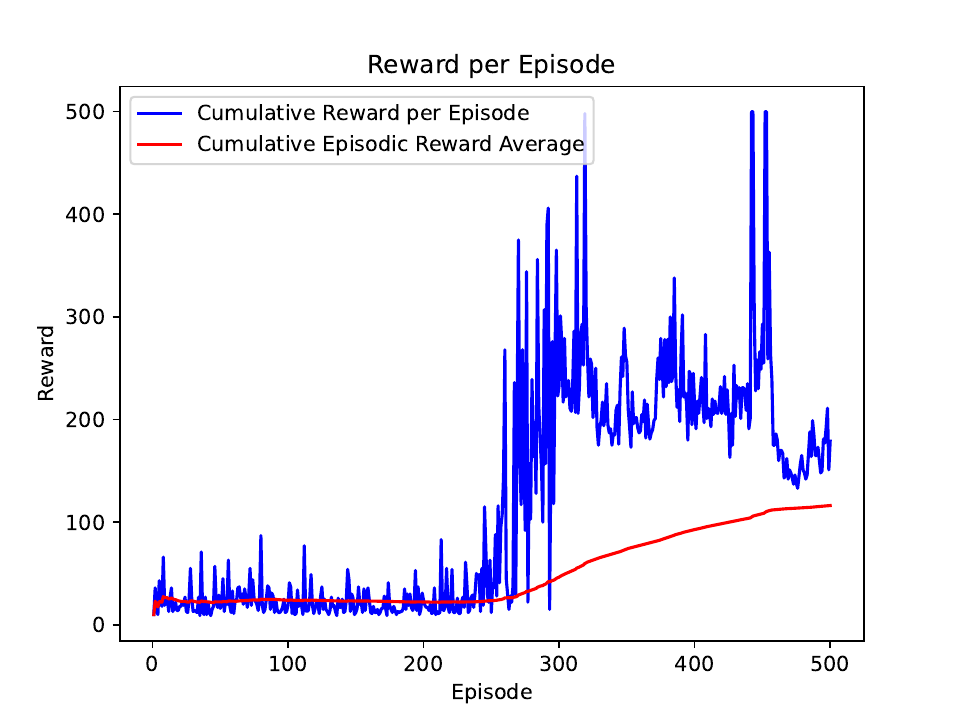}
    \caption{A graph showing the average reward as the agent goes through episodes and the overall cumulative reward per episode when employing uniform/normal experience replay. Compare to \ref{fig:perexponentialdecay}.}
    \label{fig:dqn_unif_experience}
\end{figure}

\begin{figure}[H]
    \centering
    \includegraphics[scale=0.6]{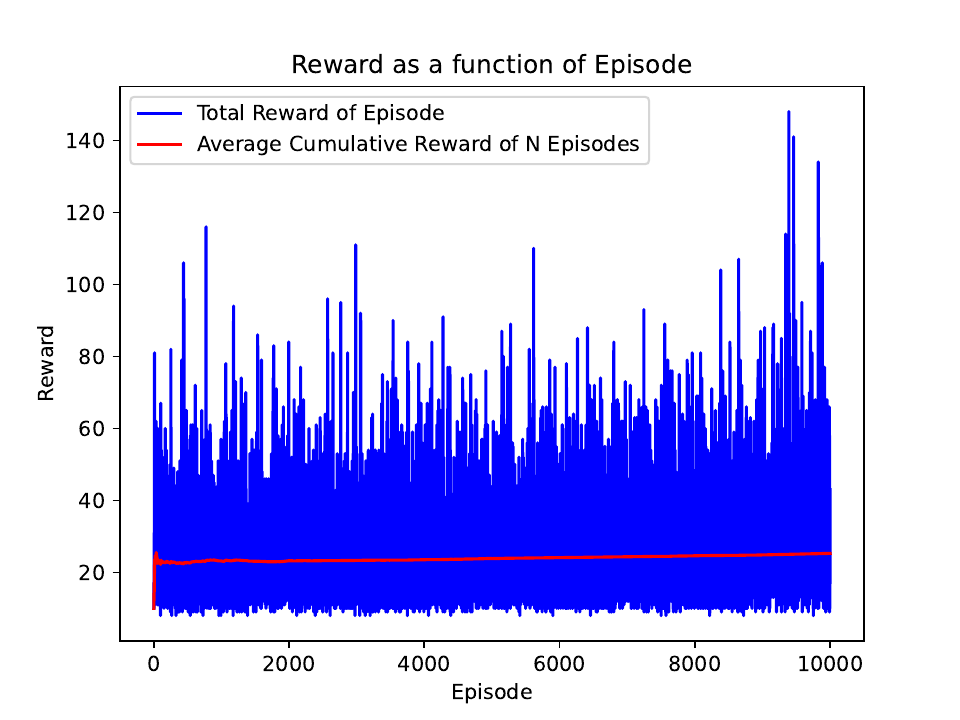}
    \caption{A graph of the reward obtained by the agent. We employed a decaying epsilon over $10,000$ episodes. Notice that the average (red) slowly increases as we use more episodes.}
    \label{fig:qlearn_10k}
\end{figure}

\end{document}